\documentclass[11pt,a4paper]{article}
\usepackage[hyperref]{emnlp2020}
\usepackage{microtype}
\usepackage{graphicx}
\usepackage{booktabs} 
\usepackage{times}
\usepackage{graphicx}
\usepackage{url} 
\usepackage{latexsym}
\usepackage{amsmath}
\usepackage{makecell}
\usepackage{float}
\usepackage{enumitem}
\usepackage{url}
\usepackage{soul}
\usepackage{multirow}
\usepackage{mathtools,kantlipsum}
\usepackage{amssymb}
\usepackage{color, colortbl}
\definecolor{Gray}{gray}{0.93}

\usepackage{microtype}

\aclfinalcopy

\newcommand{\ourModel}{\textsc{DialogRPT}}

\definecolor{lightmagenta}{rgb}{1.0, 0.42, 1.0}

\definecolor{midnightgreen}{rgb}{0.0, 0.29, 0.33}

\title{Dialogue Response Ranking\\Training with Large-Scale Human Feedback Data}

\author{Xiang Gao \quad\quad Yizhe Zhang  \quad\quad \textbf{Michel Galley}  \\ \textbf{Chris Brockett} \quad\quad \textbf{Bill Dolan}\\
  Microsoft Research, Redmond, WA, USA \\
  {\small \tt \{xiag,yizzhang,mgalley,chrisbkt,billdol\}@microsoft.com}
}
\date{}

\begin{document}
\maketitle

\begin{abstract}
Existing open-domain dialog models are generally trained to minimize the perplexity of target human responses. However, some human replies are more engaging than others, spawning more followup interactions. Current conversational models are increasingly capable of producing turns that are context-relevant, but in order to produce compelling agents, these models need to be able to predict and optimize for turns that are genuinely engaging. We leverage 
social media
feedback data (number of replies and upvotes) to build a large-scale training dataset for feedback prediction. To alleviate possible distortion between the feedback and engagingness, we convert the ranking problem to a comparison of response pairs which involve few confounding factors. We trained \ourModel, a set of GPT-2 based models on 133M pairs of human feedback data and the resulting ranker outperformed several baselines. Particularly, our ranker outperforms the conventional dialog perplexity baseline with a large margin on predicting Reddit feedback. We finally combine the feedback prediction models and a human-like scoring model to rank the machine-generated dialog responses. Crowd-sourced human evaluation shows that our ranking method correlates better with real human preferences than baseline models.\footnote{Dataset and models open-sourced on \url{https://github.com/golsun/DialogRPT}}

\end{abstract}

\section{Introduction}
\label{sec:intro}

Conversing freely in natural language is one of the greatest challenges of artificial intelligence. 
End-to-end open-domain dialog systems have become increasingly powerful, with advanced model architectures and large-scale training \cite{zhang2019dialogpt, adiwardana2020meena, roller2020blender, li2020optimus}.
In some settings, human annotators cannot reliably distinguish between human- and machine-generated responses.
Though surprisingly effective, the training objective for these models is conceptually simple: minimizing the perplexity of a reference response for a given context.

\begin{figure}
    \centering
    \includegraphics[width=0.47\textwidth]{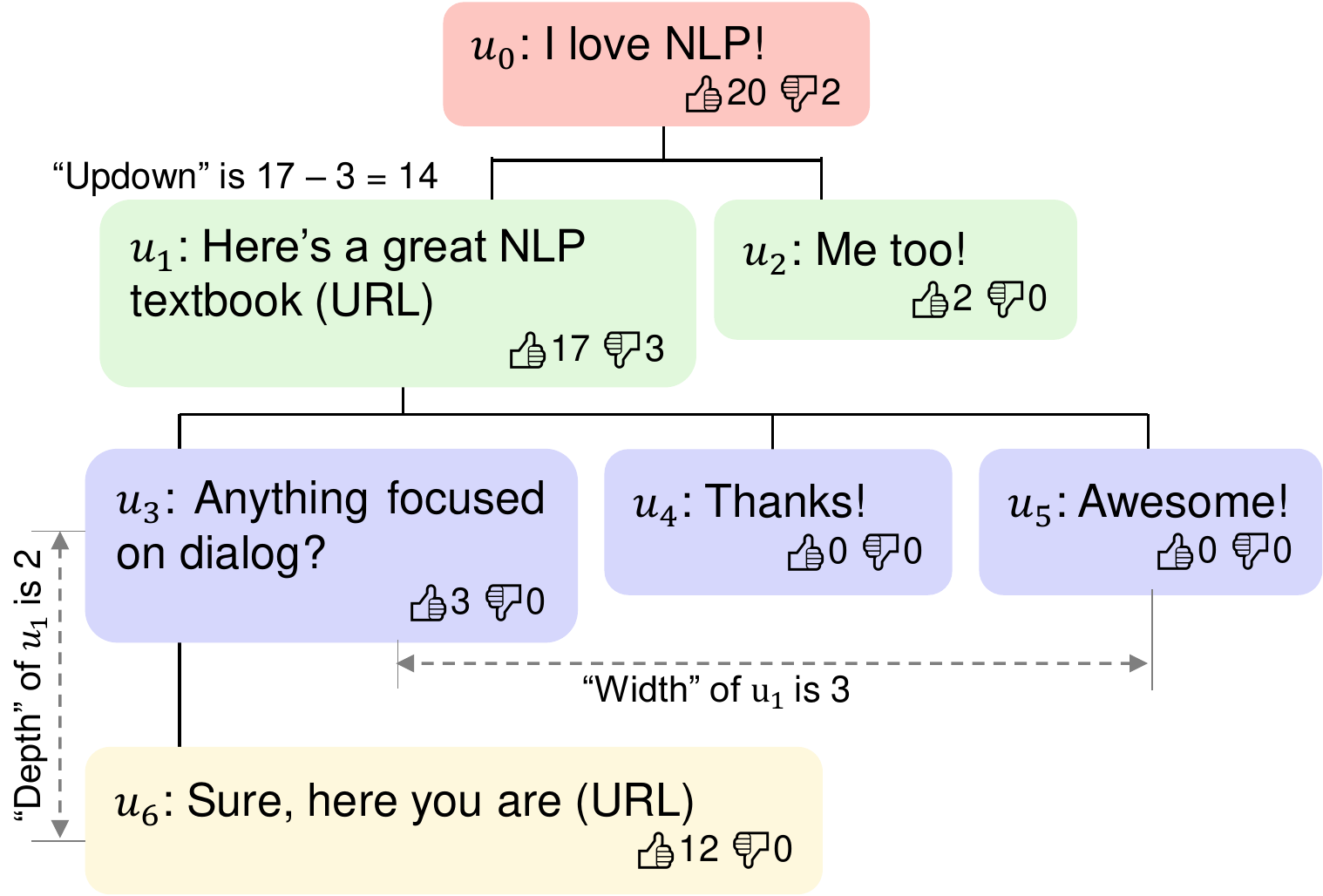}
    \caption{For many online communities, posts and comments have a tree structure and user can upvote or downvote each node individually. This allows us to define measures (e.g. Width, Depth, and Updown) of human feedback and build a large-scale training dataset for response quality prediction.}
    \label{fig:intro}
\end{figure}

However, a meaningful evaluation of response generation must take into account more than whether a generated turn is relevant in context, or whether it ``sounds human.'' Conventional neural conversation models often generate trivial or bland responses \cite{li2016mmi, zhao2017cvae} that are relevant to context but are not engaging. 
Even human responses can vary dramatically in terms of tonal appropriateness and whether they are interesting enough to prompt a rich listener reaction. A successful dialog turn must be proactive, engaging, and consistent with social norms \cite{grice1975logic, grice1989studies}.

In this work, we move beyond simple prediction of response relevance, augmenting this with a prediction of how likely a response is to elicit a positive reaction from an interlocutor. 
By incorporating a measure of engagingness into the response generation ranking algorithm, we hope to improve the overall behavior of data-driven conversational agents.

Existing methods are suboptimal for this ranking task.
Conventional perplexity based ranking methods \cite{li2016mmi, vijayakumar2016diverse} focus only on context-hypothesis relevancy.
Online conversational systems such as XiaoIce \cite{xiaoice} employ a manually-designed set of features to rank hypotheses,
but the design of these rankers is not directly based on real-world human preferences or feedback in an end-to-end fashion.
Large-scale training data is necessary because of the one-to-many nature of dialog and the scope and complexity of human conversation. However, labeling  conversations at scale is too expensive and time-consuming for this purpose. Labeling the ``engagingness'' of a response is not something a single annotator can do; the task requires something more like a large-scale, collective vote. And yet there is no obvious automated substitute for this kind of human labeling. Conventional quality measurements such as reference-based similarity \cite{papineni2002bleu} or lexical diversity \cite{li2016mmi, zhang2018gan} capture only limited aspects of response quality, and are not strongly predictive of human reactions: simply because a response is different from others does not necesarily mean that it will be perceived as ``bad''.

Our solution involves leveraging existing human feedback data (e.g., number of replies and likes) from online social communities. 
While there is work in the field of social media on feedback prediction \cite{sparling:11,stoddard2015popularity,glenski2017predicting}, 
it has not previously been applied to dialog systems and response generation. 
As illustrated in Figure~\ref{fig:intro}, each comment has its own number of replies and upvotes (termed as ``Likes” in some communities). These can be used as engagingness labels after careful normalization and formulation. There exist billions of online threads available and the number is growing fast, thus making it possible to build a large-scale training dataset. However, the relation between feedback and quality may be distorted due to social influence and other confounding factors \cite{salganik2006inequality}.

In order to ameliorate this problem, we propose a contrastive formulation, shifting from ranking to pairwise classification. 
Using a dataset of 133M pairs of human comments and their associated number of replies or up-/downvotes,
we train a set of large-scale transformer-based feedback ranking models which outperform several baselines. In particular, dialog perplexity shows little predictive power of human feedback.
We also show that a classifier trained on human-vs-artificial data can achieve good zero-shot relevancy prediction accuracy.
Finally, we describe an ensemble model that is capable of merging the predictive powers of all these models, tuned using human calibration. Human evaluation shows that our ranking method outperforms the baselines in terms of correlation with actual human preferences.

\section{Human Feedback}
Many social media platforms, such as Reddit, Twitter, and Facebook allow users to reply or upvote contents, leveraging that feedback to make decisions about what content to display, highlight, and hide. These collective ratings are treated as a proxy for content engagingness. 
In this section we discuss a few metrics of user vote data, along with some of the issues posed by its use.

\begin{table}[ht]
    \centering
    \small
    \begin{tabular}{
    p{0.09\textwidth}|p{0.09\textwidth} p{0.09\textwidth} p{0.09\textwidth}
    }
    \Xhline{2\arrayrulewidth}
           & Width  & Depth & Updown \\
    \hline
    Width 
        & \cellcolor{blue!25} 1         
        & \cellcolor{blue!21} 0.8592                
        & \cellcolor{blue!9} 0.3491 \\
    Depth 
        & \cellcolor{blue!21} 0.8592   
        & \cellcolor{blue!25} 1  
        & \cellcolor{blue!8} 0.3257 \\
    Updown 
        & \cellcolor{blue!9}  0.3491                
        & \cellcolor{blue!8}  0.3257  
        & \cellcolor{blue!25} 1\\
    \Xhline{2\arrayrulewidth}
    
    \end{tabular}
    \caption{Spearman's $\rho$ between different measurements of human feedback. Darker cell color indicates higher correlation.
    }
    \label{table:corr_label}
\end{table}

\begin{figure}[t]
    \centering
    \includegraphics[width=0.45\textwidth]{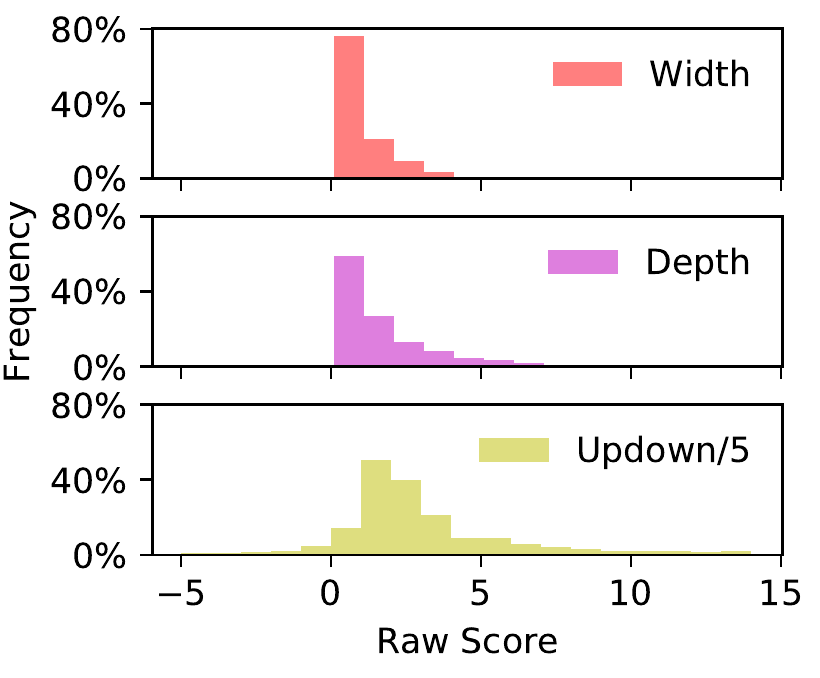}
    \caption{The long-tailed distribution of the raw scores of feedback of Reddit.com. 
    }
    \label{fig:hist}
\end{figure}

\begin{figure}[t]
    \centering
    \includegraphics[width=0.47\textwidth]{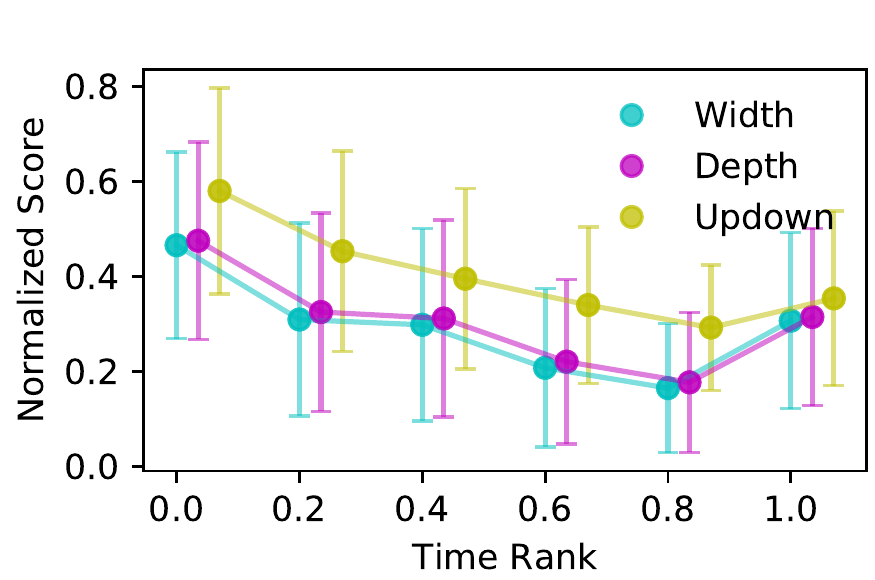}
    \caption{The dependence of feedback on created time of the comments of the same parent node (i.e. the same context) of Reddit.com. Error bars show standard deviation.}
    \label{fig:time}
\end{figure}

\subsection{Feedback metrics}

As illustrated by Figure~\ref{fig:intro}, posts and comments typically form a tree structure.
Each comment branching from the root may have its own comment children.
We consider the path from the root to the parent node of a comment to be its context $c$, and the comment as a reply $r$.
For each dialog $(c,r)$, we consider the following feedback:
\textbf{Width}, the number of direct replies to $r$; 
\textbf{Depth}, the maximum length of the dialog after this turn; and
\textbf{Updown}, the number of upvotes minus the number of downvotes. For example, given the context $c=u_0$, the reply $u_1$ gets three direct replies $u_3$, $u_4$, $u_5$ and the Width is thus 3. $u_3$ continues the dialog with one more turn $u_6$, thus the depth is 2. $u_1$ got 17 upvotes and 3 downvotes so its Updown is 14. In contrast, $u_2$ is for the same context, but its Width and Depth is only 0, and Updown is 2.

Though focused on different dimensions, both Width and Depth can be seen as measures of the number of replies, and are therefore often closely correlated, as shown in Table~\ref{table:corr_label} using Reddit as an example. They are less correlated with Updown. 
Presumably, contributors may feel that an upvote is enough to express their agreement or appreciation, and so do not post a full reply. 

\subsection{Feedback and Engagingness}

The feedback metrics defined above cannot be directly used as a measure of reply engagingness. \citet{stoddard2015popularity} shows that while popularity, measured by Updown, generally increases with quality, posts of similar quality can exhibit very different upvote counts. This variability can be traced to several different factors.
As illustrated in Figure~\ref{fig:hist}, the distribution of feedback is long-tailed, with a small fraction of threads receiving most of the replies and likes. Additionally, the popularity of the specific subreddit in which a comment occurs further confounds things: a relatively uninteresting comment in a very popular thread may get more feedback than an interesting comment in a less trafficked subreddit.
Feedback volume is also heavily dependent on the timing of a comment relative to other comments, with replies that come early in a thread being more likely to attract replies or likes. This is shown in Figure~\ref{fig:time}.
This may be tied to other factors such as social influence and disparities in comment visibility causing distortions in the relationship between comment engagingness and popularity \cite{salganik2006inequality, salganik2008astray, gilbert2013widespread}.
These findings imply that careful formulation and normalization should be applied before using feedback data as a training signal. We present our approach to this in Section~\ref{sec:pairwise}.

\subsection{Tasks}

Given a context and a list of responses, we consider the task of predicting a ranking based on the feedback they received, as measured by these three separate metrics: (1) {Width}, (2) {Depth}, and (3) {Updown}.
The gold label and training data is available for human response ranking, but in order to make this applicable to machine generated responses, we introduce another task: (4) human-vs-fake, which measures how human-like the response is. We consider two modes of fake examples: random human responses and machine generated responses. We will introduce an ensemble method in Section~\ref{sec:ensemble} for this last task.

\section{The \ourModel~Method}
In this section we introduce Dialog Ranking Pre-trained Transformers (\ourModel).

\subsection{Problem Formulation}
\label{sec:pairwise}
\paragraph{A Contrastive Learning approach.}
Given the confounding factors affecting feedback mentioned above,
we train the model on \emph{pairs} of samples $(c,r^+)$ and $(c,r^-)$, rather than fitting it to score each dialog \emph{individually}. This follows the Contrastive Learning approach (see Section~\ref{sec:related} for a brief review). The model is trained to predict a higher score for the positive sample $r^+$ (i.e. the response with more feedback) compared to the negative sample $r^-$.
Besides (1) only comparing replies of the same context, we use the following criteria to construct pairs that minimize the effect of confounding factors: 
(2) the sequence of two replies, $r^+$ and $r^-$, must have been created within a brief time window (no more than one hour), 
and (3) the feedback score of $r^+$ must exceed that of $r^-$ by a specified threshold in order to make the label less noisy. Due to the long-tailed distribution, we consider both an absolute-valued threshold and a percentage ranking threshold.
Furthermore, if a reply has more downvotes than upvotes, it will not be considered as a positive sample, but can be used as a negative sample. 

\paragraph{Training objective.}
The model should be able to output a score at testing time for a hypothesis $r$ for a given context $c$. At training time, as formulated in Section~\ref{sec:pairwise}, given two hypotheses for a context, the model should be able to identify which one has more feedback. To connect these two requirements, the model outputs a scalar $h$,
\begin{align}
    h(c,r) = \text{\ourModel}(c, r)
\end{align}
At inference time, we compute the score $s(r|c)$
\begin{align}
\label{eq:sigmoid}
    s(r|c) = \text{Sigmoid}(h(c,r))
\end{align}
For training,  the loss is designed to simultaneously maximize the positive sample score and minimize the negative sample score:
\begin{align}  
    \mathcal{L} = - \sum_{i\in \text{batch}} \log{\frac{
        e^{h(c_i, r^+_i)}
        }{
        e^{h(c_i, r^+_i)} + e^{h(c_i, r^-_i)}
        }}
\end{align}
This can be interpreted as the cross entropy between the target distribution $\{P(r^+)=1,P(r^-)=0\}$ and the predicted distribution in Softmax form.
Note the contrastive form is crucial, given that a loss function only maximizing $s(r^+|c)$ usually leads to a collapsed solution \cite{hadsell2006contrastive}.

\subsection{Model ensemble}
\label{sec:ensemble}

\paragraph{For machine generation.}
The machine generation is required to be both \textit{human-like} and \textit{preferred by human}. 
To rank the machine generations, we factorize the probability of a joint distribution as follows:
    \begin{align}\label{eq:fac}
          P(r&=\text{preferred, human-like}|c) \nonumber\\
          =& P(r=\text{preferred}| r=\text{human-like},c)\cdot \nonumber\\ 
          &P(r=\text{human-like}|c)
    \end{align} 
We estimate the first term with the models trained on a human-vs-human ranker on each feedback metric $K\in\{\text{Width, Depth, Updown}\}$
    \begin{align}
    P(r&=\text{preferred}_K, \text{human-like}|c) \triangleq s_K(r|c)
    \end{align} 
We denote the term $P(r=\text{human-like}|c)$ as $\pi_0(r|c)$, and build a classifier to predict how human-like a response is (see Section~\ref{sec:imp} for details).
    \begin{align}
    P(r&=\text{human-like}|c) \triangleq \pi_0(r|c)
    \end{align} 
Both $\pi_0(r|c)$ and $ s_K(r|c)$ are scores defined in Eq.~\ref{eq:sigmoid} interpreted as probability.

\paragraph{For overall preference.}
In case only a simple human preference matters (instead of separate Width, Depth, Updown metrics), we assume that a linear combination exists 
    \begin{align}
         s_\text{Prefer}(r|c) 
         \triangleq  \pi_0(r|c) \sum_{K} w_K s_\text{K}(r|c)  
    \label{eqn:prefer}
    \end{align}

\paragraph{Human calibration.}
To estimate the correlation between the feedback score and human response preference, we present pairs of responses for the same context to a set of human annotators, asking them to select the response they would prefer to send or receive. The annotation is conducted for machine-vs.-machine comparisons on 1K pairs, and with 5 individual judges for each pair. Through this controlled setup, we reduce confounding factors, such as social influence and disparities in visibility, that might exist even within the contrastive problem formulation. 

The results 
are used as a proxy for $s_\text{Prefer}(r|c)$, and can be used to estimate $w_K$ for the test set, though the optimal value may depend on the test set and the instructions the human annotators were given. Note that the freedom of the system is now limited to a handful of hyper-parameters, limiting the need for large-scale human labeling to learn the model parameters.

\subsection{Implementation details}
\label{sec:imp}
\paragraph{Model and training.}
Our model is a 12-layer transformer model based on GPT-2 \cite{radford2019gpt2} architecture, and initialized with DialoGPT-medium model weights \cite{zhang2019dialogpt}. DialoGPT is a large-scale dialog response generation model, pre-trained on 147M Reddit conversations.
We use a linear layer to convert the final layer transformer output at the last token time step to a scalar $h$. The parameters of the transformers and this output layer are trained simultaneously. 

Each model has 354.8M parameters, and is trained on an Nvidia Tesla V100 4-core GPU with batch size 256 at an average training speed of 0.33 M pairs of samples per hour. Each model took around 70 hours to converge (until validation loss on a fixed set of 1024 samples ceased to improve).

\begin{table}[ht]
    \centering
    \small
    \begin{tabular}{
    p{0.15\textwidth} p{0.15\textwidth}p{0.08\textwidth}
    }
    \Xhline{2\arrayrulewidth}
    \multicolumn{1}{c}{\multirow{2}{*}{Model}}
    & \multirow{2}{*}{Trained on}
    & Dataset size (M) \\
    \hline
    \multicolumn{1}{c}{\multirow{4}{*}{\shortstack{Human feedback \\$s_\text{K}(r|c)$}}}
    & Human vs. Human \\
    & - Width & 22.3\\
    & - Depth & 25.1\\
    & - Updown & 40.7\\
    \hline
    \multicolumn{1}{c}{\multirow{3}{*}{\shortstack{Human-like \\$\pi_0(r|c)$}}}
    & Human vs. Fake \\
    & - Rand & 40.7 \\
    & - Generated & 5.3 \\
    \Xhline{2\arrayrulewidth}
    
    \end{tabular}
    \caption{Summary of models and training data of different tasks, size in millions (M) of pairs 
    }
    \label{table:datasize}
\end{table}

\paragraph{Data construction.}

Following the contrastive learning approach introduced in Section~\ref{sec:pairwise}, we constructed a 133M-pair training set using Reddit data from 2011-2012, as shown in Table~\ref{table:datasize}. For each task, we sampled 1024 validation pairs from the 2012 data and 5K test pairs from the 2013 data. The train, validation and test data do not share any Reddit posts.

For the human-like (i.e. human-vs-fake) task, 
we consider two representative negative modes: retrieval and generative dialog model generation. 
For the former we simply construct negative examples by randomly sampling from the training data. For the latter we use DialoGPT with top-k decoding. Since DialoGPT is able to produce human-like responses in certain evaluation settings, we select only 5.3 M highly-rated human response as positive examples, instead of using all human responses. Note that our method can be extended to include other negative modes such as perturbations and excessive repetition, similar to the synthetic example creation using BLEURT \cite{sellam2020bleurt}.

\subsection{Baselines}

\begin{figure}[t]
    \centering
    \includegraphics[width=0.45\textwidth]{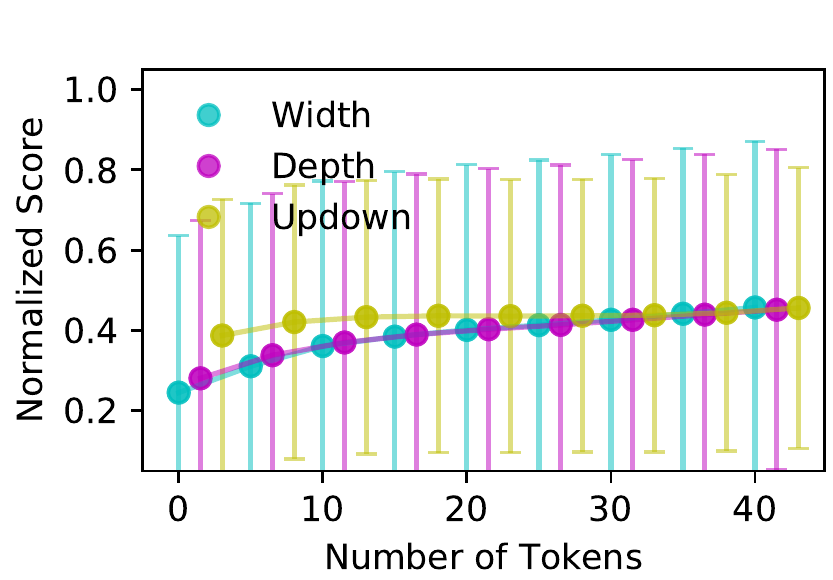}
    \caption{The dependence of feedback on the length for the comments of the same parent node (i.e. the same context). Error bars show standard deviation.}
    \label{fig:len}
\end{figure}

We consider the following baselines:
\paragraph{Dialog perplexity (ppl.)} This metric is calculated for both the forward model (i.e., predict the response from the context) and the reverse model (i.e. predict the context from the response). This ranking method was proposed by \citet{li2016mmi} and formulated to maximize mutual information (MMI) between the response and context. We use DialoGPT and its reverse model to compute ppl.

\paragraph{BM25} This classic metric measures keywords similarity \cite{robertson2009probabilistic}. We use the inner product of the context BM25 vector and candidate response BM25 vector to rank candidates, similar to \cite{Henderson2019polyai}. 

\paragraph{ConveRT} \cite{henderson2019convert} is a transformer-based model pretrained on Reddit data. It encodes context and candidate as vectors and compute their inner product as similarity used for ranking, achieved the existing state-of-the-art performance on several response matching test sets\footnote{\url{https://github.com/PolyAI-LDN/conversational-datasets/blob/master/BENCHMARKS.md}}.

\paragraph{Bag of words (BoW)} 
For each word, an average of rank-normalized feedback score\footnote{defined as $1 - i/m$, where $i$ is the feedback rank of this reply for the given context, and $m$ is the number of replies of this context} is calculated
for replies that contain this word. This is the score for this word. Due to the long-tailed distribution of the absolute value of feedback items, we normalize them as the percentage ranking for their context. Then we use the average of the scores of the words in a response as the score of this response.

\paragraph{Length} As shown in Figure~\ref{fig:len}, feedback rank weakly correlates with response length. We therefore use the average value of responses of the same length in training data as the predicted score for a hypothesis. 

~

BoW and Length baselines are are intended to capture information about lexical patterns of human feedback in the data and provide a preliminary analysis.

\section{Results}

\begin{table*}[ht]
    \centering
    \small
    \begin{tabular}{
    p{0.01\textwidth}p{0.43\textwidth}p{0.05\textwidth}p{0.05\textwidth}p{0.05\textwidth}
    }
    \Xhline{2\arrayrulewidth}
    \multicolumn{5}{c}{Context: I love NLP!}  \\
    \hline
    \multicolumn{2}{l}{Response:} & Width & Depth & Updown \\
    \emph{A} & Me too! 
        & 0.033 & 0.043 & 0.171 \\
    \emph{B} & It's super useful and more and more powerful!
        & 0.054 & 0.164 & 0.296 \\
    \emph{C} & Can you tell me how it works?
        & 0.644 & \textcolor{blue}{\textbf{0.696}} & 0.348 \\
    \emph{D} & Can anyone recommend a nice review paper?
        & \textcolor{blue}{\textbf{0.687}} & 0.562 & 0.332 \\
    \emph{E} & Here's a free textbook (URL) in case anyone needs it.
        & 0.319 & 0.409 & \textcolor{blue}{\textbf{0.612}} \\
    \Xhline{2\arrayrulewidth}
    
    \end{tabular}
    \caption{Predicted feedback scores of several example responses given the same context. 
    }
    \label{table:example}
\end{table*}

\subsection{Predicting Human Feedback}

\paragraph{Preliminary analysis}

\begin{table}[ht]
    \centering
    \small
    \begin{tabular}{
    p{0.055\textwidth} p{0.02\textwidth} p{0.28\textwidth}
    }
    \Xhline{2\arrayrulewidth}
    \multirow{4}{*}{Width} 
        & \multirow{2}{*}{\cellcolor{red!5}$r^+$}  
            & \cellcolor{red!5}url, anyone, else, who, does, why, guys, seriously, everyone\\
        & \multirow{2}{*}{\cellcolor{blue!5}$r^-$}
            & \cellcolor{blue!5}oh, amazing, damn, thanks, wow, nice, !, awesome, lol, upvote \\
    \hline
    \multirow{4}{*}{Depth}        
        & \multirow{2}{*}{\cellcolor{red!5}$r^+$}
            & \cellcolor{red!5}?, why, does, how, anyone, isn't, any, what, who,  \\
        & \multirow{2}{*}{\cellcolor{blue!5}$r^-$}
            & \cellcolor{blue!5}great, nice, amazing, damn, lol, !, awesome, thank, upvote  \\
    \hline
    \multirow{4}{*}{Updown} 
        & \multirow{2}{*}{\cellcolor{red!5}$r^+$}
            &  \cellcolor{red!5}url, our, picture, everyone, hey, part, years, into, will, we\\
        & \multirow{2}{*}{\cellcolor{blue!5}$r^-$}
            &  \cellcolor{blue!5}maybe, though, awesome, comment, funny, wow, came, upvote, lol \\
    \Xhline{2\arrayrulewidth}
    
    \end{tabular}
    \caption{Bag of words analysis. If on average the comments containing a certain word get more feedback, we list this word in the \emph{$r^+$} row. If they get less feedback, this word is listed in \emph{$r^-$} row.
    }
    \label{table:bow}
\end{table}

We first consider findings from the bag of words baseline. As shown in Table~\ref{table:bow}, responses that receive fewer replies or upvotes tend to be less contentful (e.g. \emph{lol}, \emph{awesome}, \emph{wow}, \emph{nice}). In contrast, comments that attract more feedback are typically different in character: for instance, questions (indicated by \emph{?}, \emph{why}, \emph{how}, \emph{what}, \emph{who}) often lead to longer conversation (greater Depth). Comments targeting a broad audience (labeled by \emph{anyone}, \emph{guys}), tend to receive more direct replies (greater Width) than those aimed at a specific set of people. 

A similar pattern is captured by \ourModel, as shown in Table~\ref{table:example}. Given the context \emph{I love NLP!}, the relatively bland response \emph{Me too!} gets the lowest scores for all three feedback measures. Higher scores are obtained for Response \emph{B}, where a justification is provided for the agreement (\emph{useful}, \emph{powerful}). Response \emph{C} gets the highest Depth score, as it invites a discussion about how NLP works, something that is unlikely to be completed in one or two turns. Response \emph{D}, in contrast, can be answered in fewer turns but with potentially many valid answers, which explains its high Width score. 
Finally, Response \emph{E} receives the highest Updown score, probably because the model predicts that many people will upvote it to express gratitude for the useful resource pointer it provides (\emph{textbook}). Removing the word \emph{(URL)} from Response \emph{E} causes the score to drop only slightly, indicating that the model is not simply sensitive to the post containing a web link.

\paragraph{Ranker evaluation}

\begin{table}[ht]
    \centering
    \small
    \begin{tabular}{
    p{0.05\textwidth} p{0.19\textwidth} p{0.07\textwidth} p{0.06\textwidth}
    }
    \Xhline{2\arrayrulewidth}
      & \multirow{2}{*}{Method} & Pairwise accuracy  & Spearman $\rho$ \\
    \hline
    \multirow{5}{*}{Width}
                & Dialog ppl.          & 0.513 & -0.009 \\
                & Reverse dialog ppl.  & 0.571  & 0.099  \\
                & Length baseline      & 0.595  & 0.229  \\
                & BoW baseline         & 0.596  & 0.234  \\
                & \ourModel 
                            & \textcolor{blue}{\textbf{0.752}}
                            & \textcolor{blue}{\textbf{0.357}} \\
    \hline
    \multirow{5}{*}{Depth}  
                & Dialog ppl.          & 0.508 & -0.004   \\
                & Reverse dialog ppl.      & 0.557 & 0.063 \\
                & Length baseline      & 0.543  & 0.134  \\
                & BoW baseline         & 0.584  & 0.187  \\
                & \ourModel 
                            & \textcolor{blue}{\textbf{0.695}}
                            & \textcolor{blue}{\textbf{0.317}} \\
    \hline
    \multirow{5}{*}{Updown}
                & Dialog ppl.          & 0.488 & 0.003    \\
                & Reverse dialog ppl.      & 0.560  & 0.076  \\
                & Length baseline      & 0.531 & 0.063  \\
                & BoW baseline         & 0.571  & 0.134  \\
                & \ourModel 
                            & \textcolor{blue}{\textbf{0.683}}
                            & \textcolor{blue}{\textbf{0.295}} \\
    \Xhline{2\arrayrulewidth}
    \end{tabular}
    \caption{Performance on test set ranking gold responses, measured by pairwise accuracy and Spearman's $\rho$.}
    \label{table:eval_feedback}
\end{table}

 \begin{table*}[!ht]
    \centering
    \small
    \begin{tabular}{
    p{0.12\textwidth}
    p{0.1\textwidth}
    | p{0.05\textwidth} p{0.05\textwidth} p{0.06\textwidth} 
    | p{0.05\textwidth} 
    p{0.07\textwidth}
    }   
    \Xhline{2\arrayrulewidth}
    \multicolumn{1}{c}{\multirow{3}{*}{Model}}
    & \multirow{3}{*}{Trained on}
    & \multicolumn{5}{c}{Tested on} 
            \\
            & & \multicolumn{3}{c}{Human vs. Human}  
            & \multicolumn{2}{|c}{Human vs. Fake}  
            \\
            & & Width  
            & Depth 
            & Updown 
            & Rand 
            & Generated
            \\
    \hline
    \multicolumn{1}{c}{\multirow{3}{*}{Human feedback}}
    & Width  
            & \cellcolor{blue!26} 0.764 
            & \cellcolor{blue!19} 0.693 
            & \cellcolor{blue!10} 0.601 
            & \cellcolor{blue!2}  0.517
            & \cellcolor{blue!14} 0.644
            \\
    & Depth  
            & \cellcolor{blue!25} 0.749
            & \cellcolor{blue!20} 0.701
            & \cellcolor{blue!9}  0.588
            & \cellcolor{blue!1}  0.512
            & \cellcolor{blue!15} 0.647
            \\
    & Updown 
            & \cellcolor{blue!16} 0.659     
            & \cellcolor{blue!10} 0.602       
            & \cellcolor{blue!18} 0.683 
            & \cellcolor{blue!3}  0.526
            & \cellcolor{blue!17} 0.667
            \\
    \hline
    \multicolumn{1}{c}{\multirow{2}{*}{Human-like}}
    & Rand  
            & \cellcolor{blue!6}  0.558  
            & \cellcolor{blue!5}  0.552  
            & \cellcolor{blue!2}  0.522
            & \cellcolor{blue!30} 0.843
            & \cellcolor{blue!0}  0.413
            \\
    & + Generated
            & \cellcolor{blue!8}  0.560
            & \cellcolor{blue!7}  0.558
            & \cellcolor{blue!4}  0.522
            & \cellcolor{blue!30} 0.864
            & \cellcolor{blue!30} 0.880
            \\
    \hline
    \multicolumn{1}{c}{Ensemble} & \multicolumn{1}{c}{-}
            & \cellcolor{blue!25} 0.746  
            & \cellcolor{blue!18} 0.675   
            & \cellcolor{blue!17} 0.666   
            & \cellcolor{blue!26} 0.758
            & \cellcolor{blue!30} 0.821
            \\
    \Xhline{2\arrayrulewidth}

    \end{tabular}
    \caption{Pairwise accuracy of \ourModel~models. Darker cell color indicates better performance.
    }
    \label{table:heatmap}
\end{table*}

We evaluate ranker performance using two metrics. 
First, we use pairwise accuracy, which measures accuracy in selecting the positive sample from a positive (more feedback) and negative (less feedback) pair for the same context. This is consistent with the training objective. Second, since the models will be used to rank hypotheses, we are also interested in the correlation between the model scorer rank and the the gold label rank. We measure this correlation using Spearman's $\rho$.

As shown in Table~\ref{table:eval_feedback}, \ourModel~shows the highest test performance on both measurements\footnote{Similar results are observed for the validation set.} 
Reverse dialog perplexity generally performs better than  forward dialog perplexity. 
However, as it is not trained with feedback labels, a simple BoW baseline outperforms the dialog models in this task.

We also evaluated performance on feedback data that the model had not been trained on, as shown in Table~\ref{table:heatmap}.
The model trained on Width data can perform reasonably well on Depth prediction, and vice versa, consistent with the high correlation between their labels as shown in Table~\ref{table:corr_label}. The Updown label is less correlated with these, and so the model trained on Updown data performs poorly on Width and Depth data. This is in keeping with the complementary relationship between these models.

\subsection{Human-like Classification}

\begin{table}[ht]
    \centering
    \small
    \begin{tabular}{
    p{0.11\textwidth} p{0.17\textwidth} p{0.05\textwidth}
    }
    \Xhline{2\arrayrulewidth}
    Dataset & Method  & Hits@$k$ \\
    
    \hline
    \multirow{6}{*}{\shortstack{Reddit\\($k>$5,$n$=k)}}  
        & BLEU1         & 0.651 \\
        & BERTScore     & 0.685  \\
        & BLEURT        & 0.714  \\
         \cmidrule{2-3}
        & BM25          & 0.309 \\
        & ConvRT          & 0.760 \\
        & Dialog ppl.      & 0.560  \\
        & Reverse dialog ppl.  & 0.775  \\
        & \ourModel   &  \textcolor{blue}{\textbf{0.886}}  \\
    
    \hline
    \multirow{3}{*}{\shortstack{DailyDialog\\($k$=1,$n$=19)}}
        & BM25          & 0.182 \\
        & ConvRT          & 0.380  \\
        & Dialog ppl.       & 0.176  \\
        & Reverse dialog ppl.  & 0.457  \\
        & \ourModel  &  \textcolor{blue}{\textbf{0.621}}  \\
    
    \hline
    \multirow{3}{*}{\shortstack{Twitter\\($k$=1,$n$=19)}}
        & BM25          & 0.178 \\
        & ConvRT          & 0.439  \\
        & Dialog ppl.       & 0.107  \\
        & Reverse dialog ppl.  & 0.440   \\
        & \ourModel  &  \textcolor{blue}{\textbf{0.548}}  \\
    
    \hline
    \multirow{6}{*}{\shortstack{PersonaChat\\($k$=1,$n$=19)}}
        & BM25          & 0.117 \\
        & ConvRT          & 0.197  \\
        & IR Baseline   & 0.213 \\
        & Starspace     & 0.318  \\
        & KV profile memory & 0.349  \\
        & Dialog ppl.       &  0.108 \\
        & Reverse dialog ppl.  & 0.449 \\
        & \ourModel &  \textcolor{blue}{\textbf{0.479}}  \\
    \Xhline{2\arrayrulewidth}
    \end{tabular}
    \caption{Performance ranking $k$ gold and $n$ distractor responses. {\ourModel} is trained on Reddit human-vs-rand dataset, and is \emph{zero-shot} for other datasets in the table.}
    \label{table:eval_mismatch}
\end{table}

\paragraph{Human-vs-Rand}
We first evaluate performance on the task of selecting the gold response from a set of random distractor responses. For each context, we randomly select $n$ distractors. Performance is evaluated using Hits@$k$, which is the ratio of the number of gold responses in the top-$k$ ranked hypotheses. Here, $k$ is equal to the number of gold responses. 
Although \ourModel~is trained solely on Human-vs-Rand Reddit data, we show in Table~\ref{table:eval_mismatch} that it performs well even when compared to baseline models on other data sources: DailyDialog \cite{li2017dailydialog} and Twitter\footnote{\url{https://github.com/Marsan-Ma/chat_corpus/}} PersonaChat\footnote{The performance of IR Baseline, Starspace, and KV Profile Memory for PersonaChat are following \citet{zhang2018personachat}.} \cite{zhang2018personachat}. 
Such \emph{zero-shot} performance indicate that the model generalize reasonably well on unseen datasets.

For the Reddit dataset, which has multiple gold replies, we also compare our method with reference-based similarity measurements,
\footnote{Following \citet{DSTC7task2},  for a gold hypothesis, we only use other $k-1$ gold hypotheses as references to avoid a similarity of 1. For each distractor response, we randomly pick $k-1$ references from $k$ gold hypotheses. } 
including BLEU \cite{papineni2002bleu}, BERTScore \cite{zhang2019bertscore}, and BLEURT \cite{sellam2020bleurt}. 
These metrics are not applicable on-the-fly, since references are not available, but they are commonly used as offline measures of dialog system quality. 
As shown in Table~\ref{table:eval_mismatch}, although BLEU, BERTScore, and BLEURT take advantage of reference, which is unknown to \ourModel, 
\ourModel shows higher accuracy measured by Hits@$k$.

\paragraph{Human-vs-Generated}
We evaluate the model's ability to discriminate between human and generated responses. As shown in Table~\ref{table:heatmap}, a model trained only on human-vs-rand data performs poorly on this task, indicating that the generated responses are sufficiently relevant to the context to yield a higher
score than a random response. This is consistent with the evaluation results reported by \citet{zhang2019dialogpt}, which shows that DialoGPT receives higher relevancy score in a human evaluation. However, the feedback prediction models, Width, Depth and Updown, show much higher accuracy in the human-vs-generated task, even though they were not trained on any generated responses. This implies that the ranking models predict that DialoGPT's generated responses may not be as proactive or as engaging as human responses. Finally, the model trained with both random and generated responses perform well on both human-vs.-fake tasks, but not well on the human-vs.-human feedback ranking tasks. This indicates that the models are complementary to each other, motivating us to build an ensemble model.

\subsection{Ensembling Models}

\paragraph{Reddit test data.} The feedback and the human-like 
models are combined following Eq.~\ref{eqn:prefer} and evaluated using different test sets, as shown in Table~\ref{table:heatmap}. For testing on feedback $K$, where $K$ is Width, Depth or Updown, we set $w_i=1$ if $i=K$ and 0 otherwise. For human vs. fake, we set $w_K=1/3$ for all three feedback models. Although the ensemble model's accuracy is not the highest for any of the test sets, 
it performs reasonably well on all of them.

 \begin{table}[!ht]
    \centering
    \small
    \begin{tabular}{
    p{0.15\textwidth}
    | p{0.12\textwidth} p{0.12\textwidth}
    }   
    \Xhline{2\arrayrulewidth}
            & Acc.
            & $\rho$ 
            \\
    \hline
    Dialog ppl. 
        & 0.539 (0.033) & 0.082 (0.060)  \\
    Reverse dialog ppl. 
        & 0.548 (0.031) & 0.094 (0.056)  \\
    \hline
    \ourModel \\
    ~~~~$\pi_0 s_\text{Width}$ 
        & 0.749 (0.008) & 0.465 (0.012) \\
    ~~~~$\pi_0 s_\text{Depth}$ 
        & 0.762 (0.009) & 0.467 (0.013) \\
    ~~~~$\pi_0 s_\text{Updown}$ 
        & 0.760 (0.008) & 0.470 (0.013) \\
    ~~~~$\sum_{K} w_K s_\text{K}$ 
        & 0.629 (0.014) & 0.201 (0.019)  \\
    ~~~~$\pi_0 \sum_{K} w_K s_\text{K}$ 
        & \textcolor{blue}{\textbf{0.792}} (0.010) & \textcolor{blue}{\textbf{0.518}} (0.015) \\
    
    \Xhline{2\arrayrulewidth}
    
    \end{tabular}
    \caption{Performance of human overall preference prediction measured by acurracy (Acc.) and Pearson correlation ($\rho$). 
    Values are reported in form ``average (standard error)" of 10-fold cross validation results.
    }
    \label{table:uhrs}
\end{table}

\paragraph{Human overall preference.} We also test the correlation between the ensemble model and human overall preference, using the human annotations introduced in Section~\ref{sec:ensemble}. 
As shown in Table~\ref{table:uhrs}, adding the human-like model $\pi_0$ improves the model performance, indicated by the comparison between the model $\pi_0 \sum_{K} w_K s_\text{K}$ and $\sum_{K} w_K s_\text{K}$. 
Among the three feedback modes, human preference correlates best with Updown. 
Presumably, Upvotes (or "Likes"),
is more directly tied to human preference than Width or Depth.  However, the other two metrics are useful as well.
The fitted coefficients of the $\sum_{K} w_K s_\text{K}$ model implies the overall preference is a combination of these modes, favoring replies that can prolong a dialog session ($w_\text{Depth}=0.48$), that are likely to be upvoted ($w_\text{Updown}=1.0$) and that do not target too broad an audience ($w_\text{Width}=-0.50$). 

\paragraph{Improving generation model.}

Even when the generative model (i.e. DialoGPT) is held constant, \ourModel~
improves candidate ranking in comparison to perplexity-based methods. This indicates that incorporating human feedback information into response generation ranking methods can yield improvements over methods that rely solely on measures of relevancy.

\section{Related Work}
\label{sec:related}

\paragraph{Dialog hypothesis ranking.}
Earlier work has explored the use of generation probability $P(h|x)$ or perplexity for hypothesis ranking.  
\citet{li2016mmi} combine this with reverse dialog probability to consider mutual information (MMI) in ranking dialog response hypotheses
\citet{gao2019stylefusion} adds style intensity for stylized response generation. 
Another line of works \cite{Henderson2019polyai, humeau2019fairpoly} encodes context and candidate as vectors and use their similarity for ranking.
Some systems \cite{xiaoice, gao2020mixingboard} employ a set of features to rank hypotheses, e.g., local cohesion, global coherence, empathy matching, and retrieval matching.

\paragraph{Reference-based quality measure} is also used to estimate the quality of response, although this is not applicable on-the-fly.
BLEU \cite{papineni2002bleu} is a classic metric measuring the sentence similarity using ngram overlap.
BERTScore \cite{zhang2019bertscore} uses BERT contextualized word embeddings, instead of ngrams. 
BLEURT \cite{sellam2020bleurt} directly measures sentence-level similarity, initialized with BERT and then trained on millions of synthetic examples.

\paragraph{Contrastive Learning} focuses on the relation between samples or labels. \citet{hadsell2006contrastive} learns representations using a contrastive loss function which pulls neighbors together and pushes apart non-neighbors in the learned space. 
\citet{jointly2019gao} designed a loss function to reduce the distance between matched context and response in contrast to the random pairs.
\citet{chen2020simple} proposed a contrastive learning framework, establishing a new state-of-the-art for image classification.

\paragraph{Social sciences and social-media NLP:} 
\citet{glenski2017predicting} model each user separately and predict their interaction for a given post 
using features including existing upvotes/downvotes, rank, and bag of words.
\citet{stoddard2015popularity} models upvotes as a time-series function of content quality, displaying position, age and score of the post  
and shows that popularity is positively correlated with quality, though articles of similar quality can have very different numbers of upvotes.
\citet{lakkaraju2013interplay} studied resubmissions to decompose article popularity into the quality of the content and the appeal of the title. They find that textual features of the title significantly affect popularity.

\section{Conclusion}

We leverage Reddit human feedback data to build and release a large-scale training dataset for feedback prediction.
We trained GPT-2 based models on 133M pairs of human feedback data and demonstrate that these models outperform several standard 
baselines. In particular, the conventional dialog perplexity baseline shows little predictive power on Reddit human feedback data.
We ensemble the feedback prediction models and a human-like scoring model to rank the machine generated dialog responses. Human evaluation shows that human preference is improved with our ranking method.
For the future work, we suggest to integrate the ranking models and generation model, e.g., in beam search stage or reinforcement learning using ranking score as reward signal.

\clearpage
\newpage
\bibliography{emnlp2020}

\begin{thebibliography}{34}
\expandafter\ifx\csname natexlab\endcsname\relax\def\natexlab#1{#1}\fi

\bibitem[{Adiwardana et~al.(2020)Adiwardana, Luong, So, Hall, Fiedel,
  Thoppilan, Yang, Kulshreshtha, Nemade, Lu et~al.}]{adiwardana2020meena}
Daniel Adiwardana, Minh-Thang Luong, David~R So, Jamie Hall, Noah Fiedel, Romal
  Thoppilan, Zi~Yang, Apoorv Kulshreshtha, Gaurav Nemade, Yifeng Lu, et~al.
  2020.
\newblock Towards a human-like open-domain chatbot.
\newblock \emph{arXiv preprint arXiv:2001.09977}.

\bibitem[{Chen et~al.(2020)Chen, Kornblith, Norouzi, and
  Hinton}]{chen2020simple}
Ting Chen, Simon Kornblith, Mohammad Norouzi, and Geoffrey Hinton. 2020.
\newblock A simple framework for contrastive learning of visual
  representations.
\newblock \emph{arXiv preprint arXiv:2002.05709}.

\bibitem[{Galley et~al.(2018)Galley, Brockett, Gao, Dolan, and
  Gao}]{DSTC7task2}
Michel Galley, Chris Brockett, Xiang Gao, Bill Dolan, and Jianfeng Gao. 2018.
\newblock End-to-end conversation modeling: Dstc7 task 2 description.
\newblock In \emph{DSTC7 workshop (forthcoming)}.

\bibitem[{Gao et~al.(2020)Gao, Galley, and Dolan}]{gao2020mixingboard}
Xiang Gao, Michel Galley, and Bill Dolan. 2020.
\newblock Mixingboard: a knowledgeable stylized integrated text generation
  platform.
\newblock \emph{arXiv preprint arXiv:2005.08365}.

\bibitem[{Gao et~al.(2019{\natexlab{a}})Gao, Lee, Zhang, Brockett, Galley, Gao,
  and Dolan}]{jointly2019gao}
Xiang Gao, Sungjin Lee, Yizhe Zhang, Chris Brockett, Michel Galley, Jianfeng
  Gao, and Bill Dolan. 2019{\natexlab{a}}.
\newblock Jointly optimizing diversity and relevance in neural response
  generation.
\newblock \emph{NAACL-HLT 2019}.

\bibitem[{Gao et~al.(2019{\natexlab{b}})Gao, Zhang, Lee, Galley, Brockett, Gao,
  and Dolan}]{gao2019stylefusion}
Xiang Gao, Yizhe Zhang, Sungjin Lee, Michel Galley, Chris Brockett, Jianfeng
  Gao, and Bill Dolan. 2019{\natexlab{b}}.
\newblock Structuring latent spaces for stylized response generation.
\newblock In \emph{Proc. of EMNLP}, pages 1814--1823.

\bibitem[{Gilbert(2013)}]{gilbert2013widespread}
Eric Gilbert. 2013.
\newblock Widespread underprovision on reddit.
\newblock In \emph{Proceedings of the 2013 conference on Computer supported
  cooperative work}, pages 803--808.

\bibitem[{Glenski and Weninger(2017)}]{glenski2017predicting}
Maria Glenski and Tim Weninger. 2017.
\newblock Predicting user-interactions on reddit.
\newblock In \emph{Proceedings of the 2017 IEEE/ACM International Conference on
  Advances in Social Networks Analysis and Mining 2017}, pages 609--612.

\bibitem[{Grice(1989)}]{grice1989studies}
H~Paul Grice. 1989.
\newblock \emph{Studies in the Way of Words}.
\newblock Harvard University Press.

\bibitem[{Grice(1975)}]{grice1975logic}
Herbert~P Grice. 1975.
\newblock Logic and conversation.
\newblock In \emph{Speech acts}, pages 41--58. Brill.

\bibitem[{Hadsell et~al.(2006)Hadsell, Chopra, and
  LeCun}]{hadsell2006contrastive}
Raia Hadsell, Sumit Chopra, and Yann LeCun. 2006.
\newblock Dimensionality reduction by learning an invariant mapping.
\newblock In \emph{2006 IEEE Computer Society Conference on Computer Vision and
  Pattern Recognition (CVPR'06)}, volume~2, pages 1735--1742. IEEE.

\bibitem[{Henderson et~al.(2019{\natexlab{a}})Henderson, Budzianowski,
  Casanueva, Coope, Gerz, Kumar, Mrk{\v{s}}i\'c, Spithourakis, Su, Vulic, and
  Wen}]{Henderson2019polyai}
Matthew Henderson, Pawe{\l} Budzianowski, I{\~{n}}igo Casanueva, Sam Coope,
  Daniela Gerz, Girish Kumar, Nikola Mrk{\v{s}}i\'c, Georgios Spithourakis,
  Pei-Hao Su, Ivan Vulic, and Tsung-Hsien Wen. 2019{\natexlab{a}}.
\newblock \href {https://arxiv.org/abs/1904.06472} {A repository of
  conversational datasets}.
\newblock In \emph{Proceedings of the Workshop on {NLP} for Conversational
  {AI}}.
\newblock Data available at github.com/PolyAI-LDN/conversational-datasets.

\bibitem[{Henderson et~al.(2019{\natexlab{b}})Henderson, Casanueva,
  Mrk{\v{s}}i{\'c}, Su, Vuli{\'c} et~al.}]{henderson2019convert}
Matthew Henderson, I{\~n}igo Casanueva, Nikola Mrk{\v{s}}i{\'c}, Pei-Hao Su,
  Ivan Vuli{\'c}, et~al. 2019{\natexlab{b}}.
\newblock Convert: Efficient and accurate conversational representations from
  transformers.
\newblock \emph{arXiv preprint arXiv:1911.03688}.

\bibitem[{Humeau et~al.(2019)Humeau, Shuster, Lachaux, and
  Weston}]{humeau2019fairpoly}
Samuel Humeau, Kurt Shuster, Marie-Anne Lachaux, and Jason Weston. 2019.
\newblock Poly-encoders: Transformer architectures and pre-training strategies
  for fast and accurate multi-sentence scoring.
\newblock \emph{arXiv preprint arXiv:1905.01969}.

\bibitem[{Lakkaraju et~al.(2013)Lakkaraju, McAuley, and
  Leskovec}]{lakkaraju2013interplay}
Himabindu Lakkaraju, Julian McAuley, and Jure Leskovec. 2013.
\newblock What's in a name? understanding the interplay between titles,
  content, and communities in social media.
\newblock In \emph{Seventh International AAAI Conference on Weblogs and Social
  Media}.

\bibitem[{Li et~al.(2020)Li, Gao, Li, Li, Peng, Zhang, and Gao}]{li2020optimus}
Chunyuan Li, Xiang Gao, Yuan Li, Xiujun Li, Baolin Peng, Yizhe Zhang, and
  Jianfeng Gao. 2020.
\newblock Optimus: Organizing sentences via pre-trained modeling of a latent
  space.
\newblock \emph{arXiv preprint arXiv:2004.04092}.

\bibitem[{Li et~al.(2016)Li, Galley, Brockett, Gao, and Dolan}]{li2016mmi}
Jiwei Li, Michel Galley, Chris Brockett, Jianfeng Gao, and Bill Dolan. 2016.
\newblock A diversity-promoting objective function for neural conversation
  models.
\newblock In \emph{NAACL}, pages 110--119.

\bibitem[{Li et~al.(2017)Li, Su, Shen, Li, Cao, and Niu}]{li2017dailydialog}
Yanran Li, Hui Su, Xiaoyu Shen, Wenjie Li, Ziqiang Cao, and Shuzi Niu. 2017.
\newblock Dailydialog: A manually labelled multi-turn dialogue dataset.
\newblock \emph{arXiv preprint arXiv:1710.03957}.

\bibitem[{Papineni et~al.(2002)Papineni, Roukos, Ward, and
  Zhu}]{papineni2002bleu}
Kishore Papineni, Salim Roukos, Todd Ward, and Wei-Jing Zhu. 2002.
\newblock {BLEU}: a method for automatic evaluation of machine translation.
\newblock In \emph{Proceedings of the 40th annual meeting on association for
  computational linguistics}, pages 311--318. Association for Computational
  Linguistics.

\bibitem[{Radford et~al.(2019)Radford, Wu, Child, Luan, Amodei, and
  Sutskever}]{radford2019gpt2}
Alec Radford, Jeffrey Wu, Rewon Child, David Luan, Dario Amodei, and Ilya
  Sutskever. 2019.
\newblock Language models are unsupervised multitask learners.
\newblock \emph{OpenAI Blog}, 1(8):9.

\bibitem[{Robertson and Zaragoza(2009)}]{robertson2009probabilistic}
Stephen Robertson and Hugo Zaragoza. 2009.
\newblock \emph{The probabilistic relevance framework: BM25 and beyond}.
\newblock Now Publishers Inc.

\bibitem[{Roller et~al.(2020)Roller, Dinan, Goyal, Ju, Williamson, Liu, Xu,
  Ott, Shuster, Smith et~al.}]{roller2020blender}
Stephen Roller, Emily Dinan, Naman Goyal, Da~Ju, Mary Williamson, Yinhan Liu,
  Jing Xu, Myle Ott, Kurt Shuster, Eric~M Smith, et~al. 2020.
\newblock Recipes for building an open-domain chatbot.
\newblock \emph{arXiv preprint arXiv:2004.13637}.

\bibitem[{Salganik et~al.(2006)Salganik, Dodds, and
  Watts}]{salganik2006inequality}
Matthew~J Salganik, Peter~Sheridan Dodds, and Duncan~J Watts. 2006.
\newblock Experimental study of inequality and unpredictability in an
  artificial cultural market.
\newblock \emph{science}, 311(5762):854--856.

\bibitem[{Salganik and Watts(2008)}]{salganik2008astray}
Matthew~J Salganik and Duncan~J Watts. 2008.
\newblock Leading the herd astray: An experimental study of self-fulfilling
  prophecies in an artificial cultural market.
\newblock \emph{Social psychology quarterly}, 71(4):338--355.

\bibitem[{Sellam et~al.(2020)Sellam, Das, and Parikh}]{sellam2020bleurt}
Thibault Sellam, Dipanjan Das, and Ankur~P Parikh. 2020.
\newblock Bleurt: Learning robust metrics for text generation.
\newblock \emph{Proc. of ACL}.

\bibitem[{Sparling and Sen(2011)}]{sparling:11}
E.~Isaac Sparling and Shilad Sen. 2011.
\newblock \href {https://doi.org/10.1145/2043932.2043961} {Rating: How
  difficult is it?}
\newblock In \emph{Proceedings of the Fifth ACM Conference on Recommender
  Systems}, page 149–156, New York, NY, USA. Association for Computing
  Machinery.

\bibitem[{Stoddard(2015)}]{stoddard2015popularity}
Greg Stoddard. 2015.
\newblock Popularity and quality in social news aggregators: A study of reddit
  and hacker news.
\newblock In \emph{Proceedings of the 24th international conference on world
  wide web}, pages 815--818.

\bibitem[{Vijayakumar et~al.(2016)Vijayakumar, Cogswell, Selvaraju, Sun, Lee,
  Crandall, and Batra}]{vijayakumar2016diverse}
Ashwin~K Vijayakumar, Michael Cogswell, Ramprasath~R Selvaraju, Qing Sun,
  Stefan Lee, David Crandall, and Dhruv Batra. 2016.
\newblock Diverse beam search: Decoding diverse solutions from neural sequence
  models.
\newblock \emph{arXiv preprint arXiv:1610.02424}.

\bibitem[{Zhang et~al.(2018{\natexlab{a}})Zhang, Dinan, Urbanek, Szlam, Kiela,
  and Weston}]{zhang2018personachat}
Saizheng Zhang, Emily Dinan, Jack Urbanek, Arthur Szlam, Douwe Kiela, and Jason
  Weston. 2018{\natexlab{a}}.
\newblock Personalizing dialogue agents: I have a dog, do you have pets too?
\newblock \emph{arXiv preprint arXiv:1801.07243}.

\bibitem[{Zhang et~al.(2019{\natexlab{a}})Zhang, Kishore, Wu, Weinberger, and
  Artzi}]{zhang2019bertscore}
Tianyi Zhang, Varsha Kishore, Felix Wu, Kilian~Q Weinberger, and Yoav Artzi.
  2019{\natexlab{a}}.
\newblock Bertscore: Evaluating text generation with bert.
\newblock \emph{Proc. of ICLR}.

\bibitem[{Zhang et~al.(2018{\natexlab{b}})Zhang, Galley, Gao, Gan, Li,
  Brockett, and Dolan}]{zhang2018gan}
Yizhe Zhang, Michel Galley, Jianfeng Gao, Zhe Gan, Xiujun Li, Chris Brockett,
  and Bill Dolan. 2018{\natexlab{b}}.
\newblock Generating informative and diverse conversational responses via
  adversarial information maximization.
\newblock In \emph{Advances in Neural Information Processing Systems}, pages
  1813--1823.

\bibitem[{Zhang et~al.(2019{\natexlab{b}})Zhang, Sun, Galley, Chen, Brockett,
  Gao, Gao, Liu, and Dolan}]{zhang2019dialogpt}
Yizhe Zhang, Siqi Sun, Michel Galley, Yen-Chun Chen, Chris Brockett, Xiang Gao,
  Jianfeng Gao, Jingjing Liu, and Bill Dolan. 2019{\natexlab{b}}.
\newblock Dialogpt: Large-scale generative pre-training for conversational
  response generation.
\newblock \emph{Proc. of ACL}.

\bibitem[{Zhao et~al.(2017)Zhao, Zhao, and Eskenazi}]{zhao2017cvae}
Tiancheng Zhao, Ran Zhao, and Maxine Eskenazi. 2017.
\newblock Learning discourse-level diversity for neural dialog models using
  conditional variational autoencoders.
\newblock In \emph{Proceedings of the 55th Annual Meeting of the Association
  for Computational Linguistics (Volume 1: Long Papers)}, volume~1, pages
  654--664.

\bibitem[{Zhou et~al.(2018)Zhou, Gao, Li, and Shum}]{xiaoice}
Li~Zhou, Jianfeng Gao, Di~Li, and Heung-Yeung Shum. 2018.
\newblock The design and implementation of xiaoice, an empathetic social
  chatbot.
\newblock \emph{arXiv preprint arXiv:1812.08989}.

\end{thebibliography}
\bibliographystyle{acl_natbib}

\end{document}